\documentclass[runningheads]{llncs}

\usepackage{subfigure}
\usepackage{subfigmat}
\usepackage{amsmath}
\usepackage{amssymb}

\usepackage{footnote}
\usepackage[ruled]{algorithm2e}
\usepackage{multirow}
\usepackage{color}
\usepackage{amsmath}
\usepackage{graphicx}
\usepackage{marvosym}
\usepackage{booktabs}  
\usepackage{threeparttable} 
\usepackage{array}
\usepackage{amsmath}
\usepackage{amssymb}
\usepackage{subfigure}
\usepackage{footnote}
\usepackage{epstopdf}
\usepackage{makecell}
\usepackage[ruled]{algorithm2e}
\usepackage{textcomp}
\usepackage{soul}
\usepackage{lipsum}
\usepackage{adjustbox}

\usepackage{tabularx}
\usepackage{float}
\usepackage{caption}
\usepackage{bbm}

\begin{document}
%
\title{Fine-grained Metrics for Point Cloud Semantic Segmentation}
%
%

\author{Zhuheng Lu\inst{1}\and
Ting Wu\inst{1} \and
Yuewei Dai\inst{1} \and
Weiqing Li\inst{2} \and
Zhiyong Su\inst{1}\textsuperscript{(\Letter)}}
\authorrunning{Z.Lu et al.}
%

\institute{School of Automation, Nanjing University of Science and Technology, Nanjing 210094, China\\ 
\email{ su@njust.edu.cn}\\
\and
 School of Computer Science and Engineering, Nanjing University of Science and Technology, Nanjing 210094, China\\}

\maketitle              
\begin{abstract}

Two forms of imbalances are commonly observed in point cloud semantic segmentation datasets: (1) category imbalances, where certain objects are more prevalent than others; and (2) size imbalances, where certain objects occupy more points than others.
Because of this, the majority of categories and large objects are favored in the existing evaluation metrics.
This paper suggests fine-grained mIoU and mAcc for a more thorough assessment of point cloud segmentation algorithms in order to address these issues.
Richer statistical information is provided for models and datasets by these fine-grained metrics, which also lessen the bias of current semantic segmentation metrics towards large objects.
The proposed metrics are used to train and assess various semantic segmentation algorithms on three distinct indoor and outdoor semantic segmentation datasets.

\keywords{Point cloud  \and Semantic segmentation \and Evaluation metrics.}
\end{abstract}

\section{Introduction}
Point cloud semantic segmentation algorithms aim to classify the point clouds at the point level. 
Algorithm optimization and improvement heavily depend on the choice of suitable and thorough evaluation metrics for various algorithms and application circumstances.
In the past few years, the overall accuracy (OA) is provided to evaluate the segmentation results. 
However, it is inappropriate to use only OA when the semantic segmentation datasets have long-tailed distributions, since it is biased to categories with larger number of points.
As a result, mean class accuracy (mAcc) is then used in the semantic segmentation experiment, which computes and averages the point accuracy of each category based on the datasets.
Besides, mean intersection over union (mIoU) is suggested as an evaluation metric of segmentation results, nevertheless, as the mAcc does not take false positives into account. 
Since then, dataset-level mAcc and mIoU is widely used for semantic segmentation tasks on different datasets, such as indoor scene dataset ScanNet \cite{2017ScanNet}, S3DIS \cite{20163D} and outdoor scene dataset Semantic3D \cite{hackel2017semantic3d}.

Among the existing dataset-level evaluation metrics, the prediction results of true positive, false positive, true negative and false negative are counted across all points over the whole dataset. 
These evaluation metrics still face several problems:
\begin{itemize}
\item Existing evaluation metrics exhibit a bias in favor of larger objects due to the imbalances. 
Nonetheless, a significant proportion of points are unequally distributed among categories in the majority of point cloud semantic segmentation datasets. 

\item Existing dataset-level evaluation metrics fail to capture valuable statistics about segmentation algorithms on a single point cloud or object instance, preventing comprehensive comparisons.
\end{itemize}

In this paper, fine-grained mIoU and mAcc are proposed for the evaluation of semantic segmentation algorithms.
Specifically, the point cloud-level metrics are first calculated to deal with the category imbalances. 
Then, the approximate instance-level metrics are further calculated to cope with the size imbalances.
These fine-grained metrics reduce bias toward large objects.
Furthermore, they supply a wealth of statistical information, allowing for more robust and comprehensive comparisons.

\section{Related Work}
\label{sec:Related Work}

In the semantic segmentation task, each point in the point cloud is assigned a predictive semantic label by the algorithm.
The predicted labels are compared with the manually labeled labels, and four indicators are counted for each category: true positive, false positive, false negative, and true negative. 
The most common evaluation metrics include: OA, mAcc, and mIoU.






OA represents the proportion of correctly segmented points to the total number of points \cite {nguyen20133d,wang2019dynamic}.
However, overall accuracy is biased to categories with large number of points.
Moreover, the overall accuracy contains limited information, making it easy to overlook the phenomenon of poor segmentation performance of a certain category.
It cannot accurately report the segmentation results of particular categories.

mAcc is introduced to average point-wise accuracy across all categories \cite{nguyen20133d,wang2019dynamic}.
It is applicable to datasets with a large and imbalanced number of categories, and can represent the semantic segmentation accuracy of each category.
However, mAcc does not take false positives in account, which leads to over-segmentation of the results.

IoU, also known as the Jaccard Similarity Coefficient (JSC), is now the most commonly employed in semantic segmentation \cite{nguyen20133d,qi2017pointnet}.
In point cloud semantic segmentation, IoU represents the overlap rate between the predicted mask and the labeled point cloud, which is used to evaluate the accuracy of the predicted segmentation area. 
It is also used in object detection and instance segmentation.
Mean intersection over union (mIoU) is the arithmetic mean of the IoU values of all categories, used to evaluate point overlap in the overall dataset.
As the most widely used metric in point cloud semantic segmentation, the mIoU indicates the prediction ability of the segmentation methods to the correct point in the datasets.

Therefore, selecting appropriate and comprehensive evaluation metrics for different algorithms plays an important role in algorithm optimization and improvement. 
Traditional dataset-level evaluation metrics cannot capture valuable statistical information regarding the performance of segmentation methods on a single point cloud or instance, preventing comprehensive comparisons.

\section{Fine-grained mIoU}
\label{sec:iou}

This section first reviews the dataset-level $\rm{mIoU^{D}}$, and then proposes point cloud-level $ \rm{mIoU^{P}}$, $ \rm{mIoU^{C}}$ and instance-level $\rm mIoU^{I}$.

\subsection{Dataset-level $\rm{mIoU^{D}}$}

The dataset-level $\rm{mIoU^{D}}$ is calculated by accumulating the segmentation results of all points in the entire dataset. 
For each category, the ${\rm IoU}_{c}^{\rm{D}}$ is defined as:
\begin{equation}\label{eq.ioud}
 {\rm IoU}_{c}^{\rm{D}}=\frac{\displaystyle\sum_{p=1}^{P} {\rm TP}_{p,c}}{\displaystyle\sum_{p=1}^{P}({\rm TP}_{p,c}+{\rm FP}_{p,c}+{\rm FN}_{p,c})},
\end{equation}
where $P$ is the number of point clouds, $\rm TP_{p,c}$, $\rm  FP_{p,c}$ and $\rm FN_{p,c}$ respectively represent the number of TP, FP, and FN points in the $c$-th category of $p$-th point cloud. 
Then, dataset-level $\rm mIoU^{D}$ is defined as:
\begin{equation}\label{eq.ioud}
 {\rm mIoU^{D}}=\frac{1}{C}\displaystyle\sum_{c=1}^{C}{\rm IoU}_{c}^{\rm{D}},
\end{equation}
where $C$ is the number of categories in the dataset.

\subsection{Point Cloud-level $\rm{mIoU^{P}}$ and $\rm{mIoU^{C}}$}
The point cloud-level metrics aim to give more detailed assessment for semantic segmentation methods to cope with the category imbalances. 
Since not all categories appear in different point cloud models, there will be statistical bias in the statistical results of the dataset-level evaluation metrics.
Therefore, the NULL representation is introduced into the point cloud-level $\rm{mIoU^{P}}$ and $\rm{mIoU^{C}}$.
For the $p$-th point cloud in the $c$-th category, the ${\rm IoU}_{p,c}$ is defined as:
\begin{equation}
 {\rm IoU}_{p,c}=\frac{{\rm TP}_{p,c}}{{\rm TP}_{p,c}+{\rm FP}_{p,c}+{\rm FN}_{p,c}}.
\end{equation}

Then, based on the ${\rm IoU}_{p,c}$ of all categories, the ${\rm IoU}_{p}^{{\rm P}}$ of $p$-th point cloud is defined as follows:
\begin{equation}
{\rm IoU}_{p}^{{\rm P}}=\frac{\displaystyle\sum_{c=1}^{C} \mathbbm{1}\{{\rm IoU}_{p,c}\neq \mathbf{NULL}\}{\rm IoU}_{p,c}}{\displaystyle\sum_{c=1}^{C} \mathbbm{1}\{{\rm IoU}_{p,c}\neq \mathbf{NULL}\}},
\end{equation}
where $\mathbbm{1}$ is the indicator function, and only non-empty categories are counted. 
If a category does not appear in the point cloud, it is marked as NULL.

Finally, $ {\rm IoU}_{p}^{{\rm P}}$ is obtained by calculating the average value of each point cloud-level ${\rm IoU}_{p}^{{\rm P}}$:
\begin{equation}\label{eq.ioud}
 {\rm mIoU^{P}}=\frac{1}{P}\displaystyle\sum_{p=1}^{P}{\rm IoU}_{p}^{{\rm P}}.
\end{equation}

\begin{figure}[]
	\centering
	\includegraphics[scale=0.70]{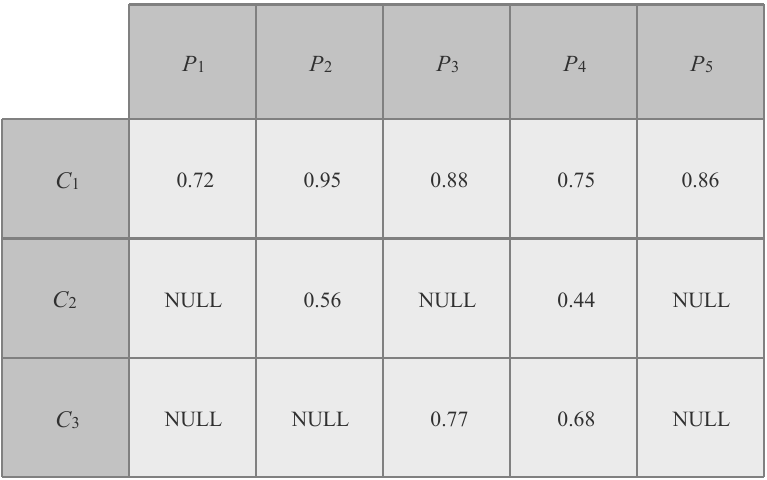}\\
	\caption{Statistical chart of mIoU values.}
	\label{fig:NULL}
\end{figure}

The results of calculating the average by category and then by point cloud are different from those of calculating the average by point cloud and then by category due to the NULL values, as shown in Figure \ref{fig:NULL}. One drawback of the $\rm mIoU^{P}$ metric is that it tends to favor category $C_{1}$, which appears more frequently in point clouds. 
To address this bias, the $\rm mIoU^{C}$ metric is first counted by point cloud and then by category. 
The point cloud-level ${\rm IoU}_{c}^{{\rm C}}$ of a certain category is defined as:
\begin{equation}
{\rm IoU}_{c}^{{\rm C}}=\frac{\displaystyle\sum_{p=1}^{P} \mathbbm{1}\{{\rm IoU}_{p,c}\neq \mathbf{NULL}\}{\rm IoU}_{p,c}}{\displaystyle\sum_{p=1}^{P} \mathbbm{1}\{{\rm IoU}_{p,c}\neq \mathbf{NULL}\}}.
\end{equation}

Similar to $\rm mIoU^{P}$, averaging these point cloud-level metrics can yield $\rm mIoU^{C}$:
\begin{equation}\label{eq.ioud}
 {\rm mIoU^{C}}=\frac{1}{C}\displaystyle\sum_{c=1}^{C}{\rm IoU}_{c}^{{\rm C}}.
\end{equation}

\subsection{Instance-level $\rm{mIoU^{I}}$}
The instance-level $\rm{mIoU^{I}}$ provides a more precise evaluation of segmentation algorithms to deal with size-variance. 
Specifically, for the $i$-th instances of $c$-th category in $p$-th point cloud, both TP and FN can be accurately calculated. 
The instance-level FP is approximated by allocating point cloud-level FP proportionally to the size of each instance \cite{wang2024revisiting}.
Then, the ${\rm IoU}_{p,c,i}$ of a a certain instance is defined as:
\begin{equation}
 {\rm IoU}_{p,c,i}=\frac{ {\rm TP}_{p,c,i}}{({\rm TP}_{p,c,i}+{\rm FN}_{p,c,i}+\frac{S_{p,c,i}}{\displaystyle\sum_{i=1}^{I_{p,c}}S_{p,c,i}}{\rm FP}_{p,c})},
\end{equation}
where $ {\rm FP}_{p,c}$ is the number of points of the $c$-th category in the $p$-th point cloud where the segmentation result is FP, and $I_{p,c}$ represents the total number of instances belonging to the $c$-th category in the $p$-th point cloud. 
$S_{p,c,i}$ is the total number of points of the $i$-th instance, which can be defined as:
\begin{equation}
S_{p,c,i}={\rm TP}_{p,c,i}+{\rm FN}_{p,c,i}.
\end{equation}

Based on the ${\rm IoU}_{p,c,i}$ of each instance, the ${\rm IoU}_{c}^{{\rm I}}$ of a certain category can be defined as follows:
\begin{equation}
{\rm IoU}_{c}^{{\rm I}}=\frac{\displaystyle\sum_{p=1}^{P}\displaystyle\sum_{i=1}^{I_{p,c}}{\rm IoU}_{p,c,i}}{\displaystyle\sum_{p=1}^{P}I_{p,c}}.
\end{equation}

By averaging ${\rm IoU}_{c}^{{\rm I}}$ of all categories, the instance-level $\rm mIoU^{I}$ metric can be obtained as follows:
\begin{equation}
{\rm mIoU^{I}}=\frac{1}{C}\displaystyle\sum_{c=1}^{C}{\rm IoU}_{c}^{{\rm I}}.
\end{equation}

\section{Fine-grained mAcc}
\label{sec:acc}

This section first reviews the dataset-level $\rm{mAcc^{D}}$, and then presents point cloud-level $ \rm{mAcc^{P}}$, $ \rm{mAcc^{C}}$ and instance-level $\rm mAcc^{I}$.

\subsection{Dataset-level $\rm{mAcc^{D}}$}
$\rm{mAcc^{D}}$ reports the average segmentation accuracy of various categories in the dataset. 
For the $c$-th category, the ${\rm Acc}_{c}^{\rm{D}}$ is defined as:
\begin{equation}\label{eq.ioud}
 {\rm Acc}_{c}^{\rm{D}}=\frac{\displaystyle\sum_{p=1}^{P}({\rm TP}_{p,c}+{\rm TN}_{p,c})}{\displaystyle\sum_{p=1}^{P}({\rm TP}_{p,c}+{\rm FP}_{p,c}+{\rm TN}_{p,c}+{\rm FN}_{p,c})}.
\end{equation}

Then, dataset-level $\rm mAcc^{D}$ is defined as:
\begin{equation}\label{eq.ioud}
 {\rm mAcc^{D}}=\frac{1}{C}\displaystyle\sum_{c=1}^{C}{\rm Acc}_{c}^{\rm{D}}.
\end{equation}

\subsection{Point Cloud-level $\rm{mAcc^{P}}$ and $\rm{mAcc^{C}}$}
The point cloud-level $\rm{mAcc^{P}}$ and $\rm{mAcc^{C}}$ are designed to provide a more comprehensive evaluation of segmentation algorithms with a focus on category imbalance.
The NULL representation is also introduced into the point cloud-level $\rm{mAcc^{P}}$ and $\rm{mAcc^{C}}$ similar to the point cloud-level $\rm{mIoU^{P}}$ and $\rm{mIoU^{C}}$.
For the $p$-th point cloud in the $c$-th category, the ${\rm Acc}_{p,c}$ is defined as:
\begin{equation}
{\rm Acc}_{p,c}=\frac{{\rm TP}_{p,c}+{\rm TN}_{p,c}}{{\rm TP}_{p,c}+{\rm FP}_{p,c}+{\rm FN}_{p,c}+{\rm TN}_{p,c}}.
\end{equation}

The ${\rm Acc}_{p}^{{\rm P}}$ for $p$-th point cloud can be calculated using the ${\rm Acc}_{p,c}$ of all categories, which is specified as:
\begin{equation}
{\rm Acc}_{p}^{{\rm P}}=\frac{\displaystyle\sum_{c=1}^{C} \mathbbm{1}\{{\rm Acc}_{p,c}\neq \mathbf{NULL}\}{\rm Acc}_{p,c}}{\displaystyle\sum_{c=1}^{C} \mathbbm{1}\{{\rm Acc}_{p,c}\neq \mathbf{NULL}\}},
\end{equation}
where $\mathbbm{1}$ is the indicator function. Only non-empty categories are counted in the ${\rm Acc}_{p}^{{\rm P}}$. 
Then, the point cloud-level $\rm mAcc^{P}$ is defined as:
\begin{equation}\label{eq.ioud}
{\rm mAcc^{P}}=\frac{1}{P}\displaystyle\sum_{p=1}^{P}{\rm Acc}_{p}^{{\rm P}}.
\end{equation}

Then, segmentation results are counted by point cloud and then by category. 
Specifically, the point cloud level ${\rm Acc}_{c}^{{\rm C}}$ of a certain category is defined as:
\begin{equation}
{\rm Acc}_{c}^{{\rm C}}=\frac{\displaystyle\sum_{p=1}^{P} \mathbbm{1}\{{\rm Acc}_{p,c}\neq \mathbf{NULL}\}{\rm Acc}_{p,c}}{\displaystyle\sum_{p=1}^{P} \mathbbm{1}\{{\rm Acc}_{p,c}\neq \mathbf{NULL}\}}.
\end{equation}

The point cloud-level ${\rm mAcc^{C}}$ is achieved by averaging all the ${\rm Acc}_{c}^{{\rm C}} $:
\begin{equation}\label{eq.ioud}
{\rm mAcc^{C}}=\frac{1}{C}\displaystyle\sum_{c=1}^{C}{\rm Acc}_{c}^{{\rm C}}.
\end{equation}

\subsection{ Instance-level $\rm{mAcc^{I}}$}
The instance-level $\rm mAcc^{I}$ is designed to provide a more granular evaluation for segmentation algorithms to deal with size differences among objects.
For the $i$-th instances of $c$-th category in $p$-th point cloud, the ${\rm Acc}_{p,c,i}$ is defined as:
\begin{equation}
{\rm Acc}_{p,c,i}=\frac{ {\rm TP}_{p,c,i}+{\rm TN}_{p,c,i}}{({\rm TP}_{p,c,i}+{\rm FN}_{p,c,i}+{\rm TN}_{p,c,i}+\frac{S_{p,c,i}}{\displaystyle\sum_{i=1}^{I_{p,c}}S_{p,c,i}}{\rm FP}_{p,c})}.
\end{equation}

Then, the ${\rm Acc}_{c}^{{\rm I}}$ of a certain category can be defined as follows:
\begin{equation}
{\rm Acc}_{c}^{{\rm I}}=\frac{\displaystyle\sum_{p=1}^{P}\displaystyle\sum_{i=1}^{I_{p,c}}{\rm Acc}_{p,c,i}}{\displaystyle\sum_{p=1}^{P}I_{p,c}}.
\end{equation}

Finall, the instance-level $\rm mAcc^{I}$ is defined as:
\begin{equation}
{\rm mAcc^{I}}=\frac{1}{C}\displaystyle\sum_{c=1}^{C}{\rm Acc}_{c}^{{\rm I}}.
\end{equation}

\section{Experimental Results}
\label{experient}
The proposed fine-grained metrics are mainly evaluated on two large-scale indoor datasets and one challenging outdoor dataset, and the benchmark results are reported for each dataset.

\subsection{Datasets}

\textbf{ScanNet}. ScanNet \cite{2017ScanNet} is a large 3D dataset containing 1,513 point clouds of scans from 707 unique indoor scenes. ScanNet contains a variety of spaces such as offices, apartments, and bathrooms. The annotation of the point clouds corresponds to 20 semantic categories plus one for the unannotated space. We adopt the official train-val split, where there are 1205 training scenes and 312 validation scenes.

\textbf{S3DIS}. S3DIS \cite{20163D} is produced for indoor scene understanding and is widely used in the point cloud semantic segmentation task. S3DIS consists of 3D RGB point clouds of six floors from three different buildings split into individual rooms. Each room is scanned with RGBD sensors and is represented by a point cloud with coordinate information and RGB value. 
It contains 3D scans of 271 rooms with 13 categories, where area 5 is used for testing and the rest for training.  

\textbf{Semantic3D}. Semantic3D \cite{hackel2017semantic3d} is a classic outdoor dataset containing urban and rural scenes. The annotation of the point clouds corresponds to 8 semantic categories, including man-made terrain, natural terrain, high vegetation, low vegetation, buildings, hard scape, scanning artifacts and cars. Semantic3D consists of 15 training scans and 15 test scans.

\begin{table}[htbp!]

	\caption{Comparison of segmentation results for different levels of mIoU metrics on the ScanNet.}
	\centering

	\begin{tabular}{lcccc}
		\hline
		Method & $\rm mIoU^{D} $ (\%)&         $\rm mIoU^{P}(\%) $      &  $\rm mIoU^{C}$ (\%) &         $\rm mIoU^{I}(\%) $ \\ \hline
	
		PointNet++ \cite{qi2017pointnet++}     &         33.9                               &    46.6& 33.1 &  32.5  \\
		PointCNN \cite{li2018pointcnn}   &           45.8                             &  58.1   &43.5&42.2    \\
		DGCNN \cite{wang2019dynamic}    &            56.3                            &      68.1  &62.3& 50.3 \\
		 KPConv \cite{thomas2019kpconv}   &        68.4                                &    72.1   &66.3& 63.5  \\
		SparseConvNet \cite{choy20194d}    &        73.6                                &      79.2  &71.7& 69.8 \\ 
		 VMNet \cite{liu2019densepoint}   &       74.6                                 &      80.4  &71.6& 69.2 \\ 
		 ConvNet+CBL \cite{zhao2021point}   &       \textbf{76.6}                                 &    81.2   &72.0& 71.2   \\ 
   PointTransformerV2 \cite{wu2022point}   &       75.2                                 &    80.7   & 72.5 & \textbf{71.8} \\ 
   OctFormer \cite{wang2023octformer}   &       76.5                                 &    \textbf{81.4}   &\textbf{72.6}& 71.7 \\ \hline
		
	\end{tabular}
	\label{tab:ScanNetiou}
\end{table}

\begin{table}[h!]

	\caption{Comparison of segmentation results for different levels of mIoU metrics on the S3DIS.}
	\centering
 
	\begin{tabular}{lcccc}
		\hline
		Method & $\rm mIoU^{D} $ (\%)&         $\rm mIoU^{P}(\%) $      &  $\rm mIoU^{C}$ (\%) &  $\rm mIoU^{I}$ (\%)\\ \hline
		PointNet \cite{qi2017pointnet}   &     30.7       &    48.7  &30.3&   28.6\\
		
		PointCNN \cite{li2018pointcnn}   &       57.3                                 &   67.7    &55.0& 53.9  \\
		DGCNN \cite{wang2019dynamic}    &        60.4                                &      70.3   &57.9& 55.7\\
		SPGraph \cite{2018Large}  &          58.0                              &    69.3   &56.8&  55.9 \\
		3D RNN \cite{ye20183d}   &              53.4                          &      65.2  &50.9& 49.8 \\ 
		 JSNet \cite{zhao2020jsnet}   &         54.5                               &      67.0   &52.6& 51.0\\ 
		 VMNet \cite{liu2019densepoint}    &        57.8                                &   67.4    &55.1& 53.4\\
		ConvNet+CBL \cite{zhao2021point}   &           69.4                             &    75.9   &66.9& 65.2 \\ 
  PointMeta \cite{lin2023meta}   &          \textbf{77.0}                              &    \textbf{81.9}   &\textbf{73.2}& \textbf{71.2\textbf} \\
  Superpoint Transformer \cite{robert2023efficient}   &           76.0                            &    80.6   &72.9& 70.1 \\ \hline
		
	\end{tabular}
	\label{tab:S3DISiou}
\end{table}

\begin{table}[htbp!]

	\caption{Comparison of segmentation results for different levels of mIoU metrics on the Semantic3D.}
	\centering
 
	\begin{tabular}{lcccc}
		\hline
		Method & $\rm mIoU^{D} $ (\%)&         $\rm mIoU^{P}(\%) $      &  $\rm mIoU^{C}$ (\%) &  $\rm mIoU^{I}$ (\%)\\ \hline
		
		TMLC-MSR \cite{hackel2016fast}    &      49.4                                  &      65.1 &46.5&  44.1 \\
		PointNet++ \cite{qi2017pointnet++}  &      63.1                                  &    76.3   &60.3&  58.1 \\
		SnapNet \cite{boulch2018snapnet}   &        59.1                                &      74.2 &55.4& 54.0 \\ 
		
		SPGraph \cite{2018Large}    &           76.2                             &    81.5  &71.5&  68.9 \\
		ConvNet+CBL \cite{zhao2021point}    &     75.0                                   &  80.6  &70.5&  68.1   \\ 
   	ConvPoint \cite{BOULCH202024}   &       76.5                                 &    81.4   &72.2& \textbf{71.7} \\
    	RandLA-Net \cite{hu2020randla}   &       77.4                                 &    \textbf{83.0}   &\textbf{74.1}& 71.0 \\
SCF-Net \cite{fan2021scf}   &       \textbf{77.6}                                 &    82.7   &73.2& 71.2 \\
  \hline
		
	\end{tabular}
	\label{tab:Semantic3Diou}
\end{table}

\subsection{Evaluations on mIoU}

This section calculates the semantic segmentation results of dataset-based metric $\rm mIoU^{D} $, point cloud-level metrics $\rm mIoU^{P}$ and $\rm mIoU^{C}$, and instance-level metric $\rm mIoU^{I}$ on three datasets. 
The evaluation results are shown in Table \ref{tab:ScanNetiou}, \ref{tab:S3DISiou}, and \ref{tab:Semantic3Diou}, respectively. 
On the ScanNet and Semantic3D datasets, none of the methods performed optimally on all of the evaluation metrics. 
Thus it is necessary to evaluate multiple evaluation metrics at the same time in order to conduct a comprehensive evaluation.
On the S3DIS dataset, the PointMeta \cite{lin2023meta} outperforms other methods on all metrics.
The segmentation results on the point cloud-level $\rm mIoU^{C}$ and the instance-level $\rm mIoU^{I}$ are close on the three datasets.
Therefore, when the dataset has no instance-level labels, the segmentation result of the point cloud-level $\rm mIoU^{C}$ can be utilized as a reference.

\begin{figure}[htbp!]
\centering
 \begin{subfigmatrix}{3}                 
  \subfigure[$\rm mIoU^{D}$ and $\rm mIoU^{C}$]{\includegraphics[width=0.31\textwidth]{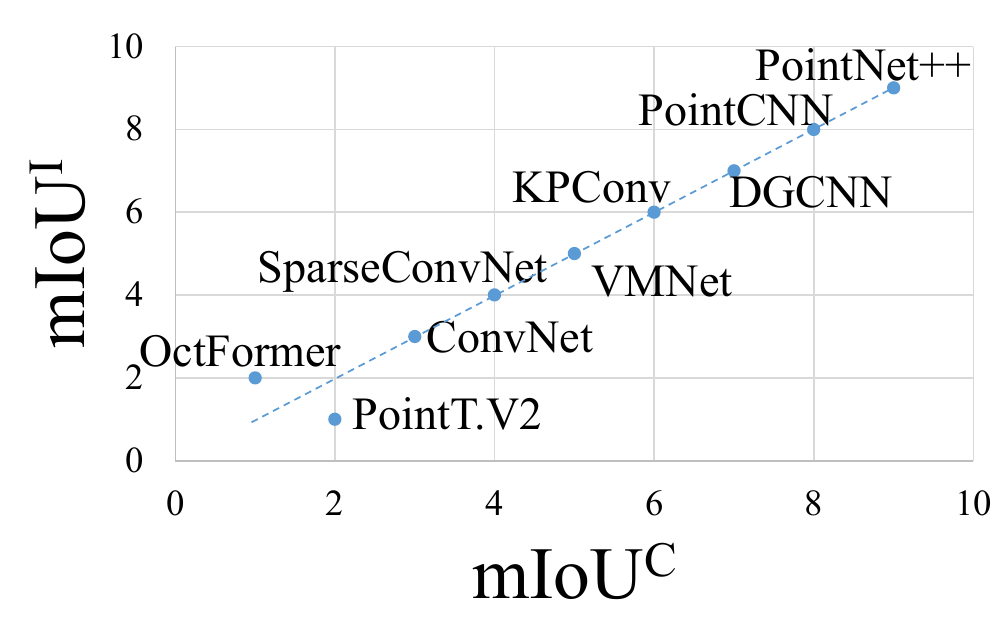}
  \label{fig:subfig:scannetiou}}
  \subfigure[$\rm mIoU^{P}$ and $\rm mIoU^{C}$]{\includegraphics[width=0.31\textwidth]{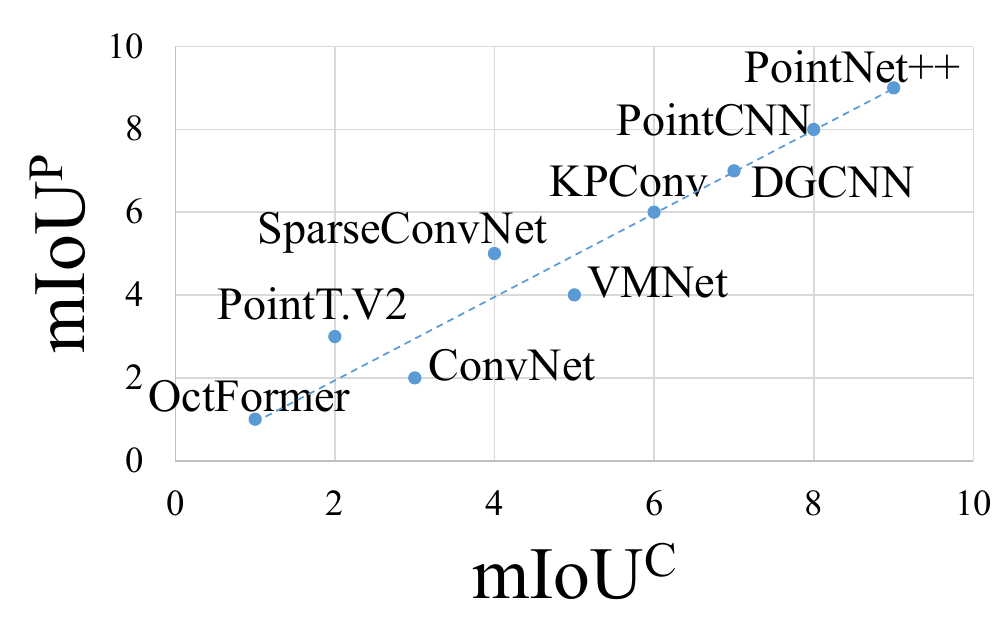}
  \label{fig:subfig:scannetioup}}
   \subfigure[$\rm mIoU^{I}$ and $\rm mIoU^{C}$]{\includegraphics[width=0.31\textwidth]{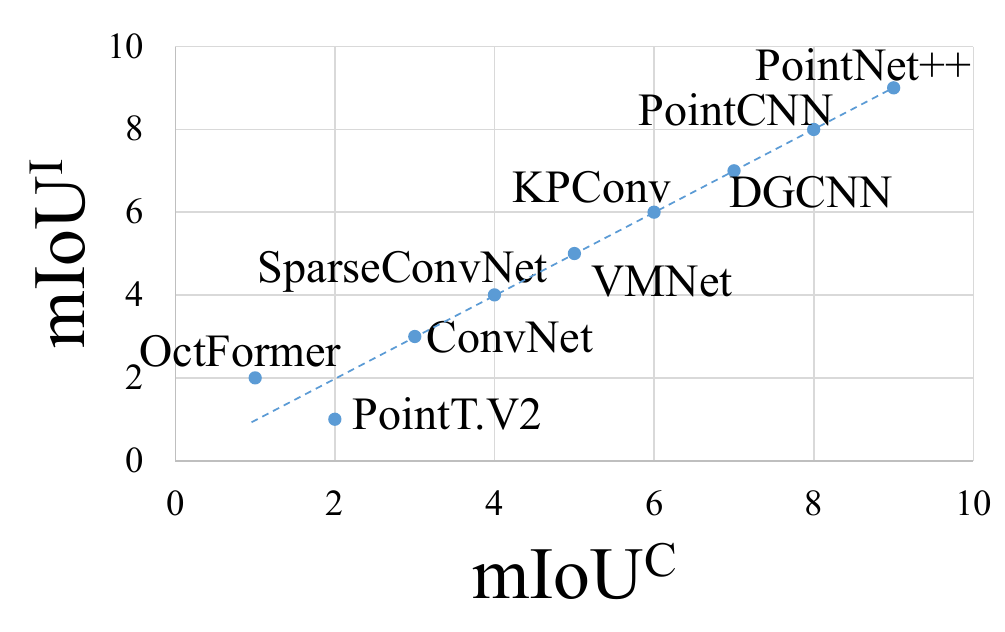}
  \label{fig:subfig:scannetioui}}
 \end{subfigmatrix}
 \caption{Comparing the rank of $\rm mIoU^{C}$ with $\rm mIoU^{D}$, $\rm mIoU^{P}$ and $\rm mIoU^{I}$ on the ScanNet.}
 \label{fig:rankiou}
\end{figure}

\begin{figure}[htbp!]
\centering
 \begin{subfigmatrix}{3}                 
  \subfigure[$\rm mIoU^{D}$ and $\rm mIoU^{C}$]{\includegraphics[width=0.31\textwidth]{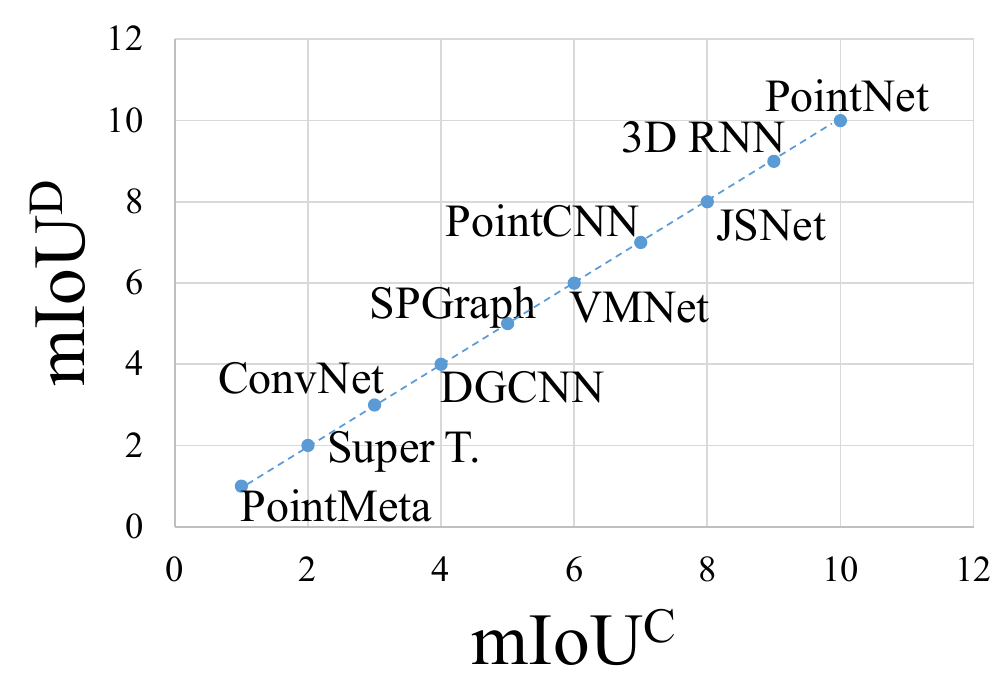}
  \label{fig:subfig:S3DISiou}}
  \subfigure[$\rm mIoU^{P}$ and $\rm mIoU^{C}$]{\includegraphics[width=0.31\textwidth]{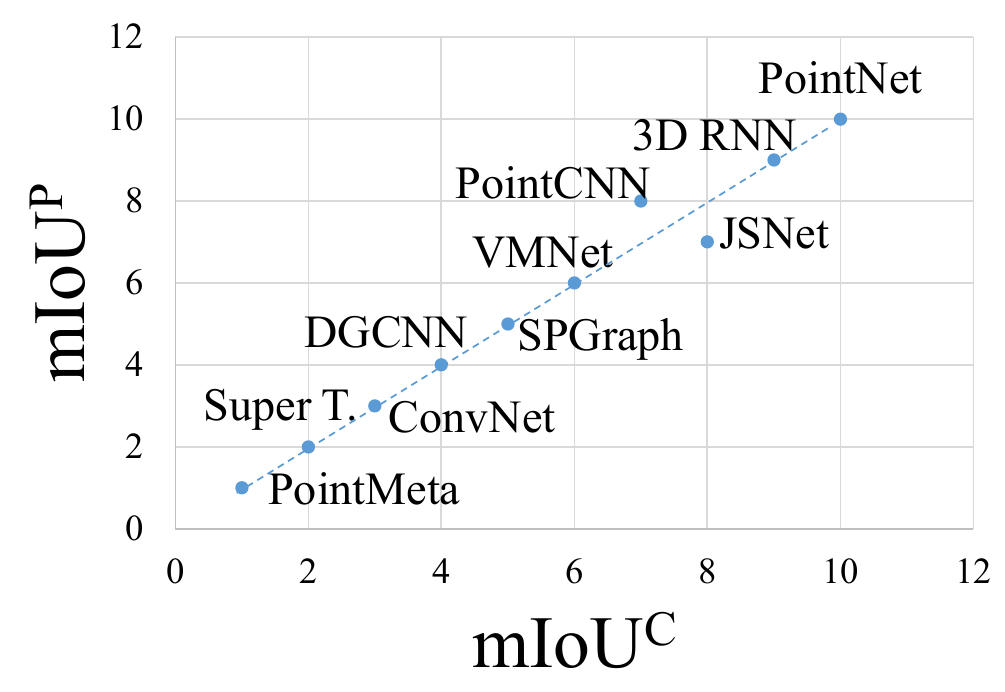}
  \label{fig:subfig:S3DISioup}}
   \subfigure[$\rm mIoU^{I}$ and $\rm mIoU^{C}$]{\includegraphics[width=0.31\textwidth]{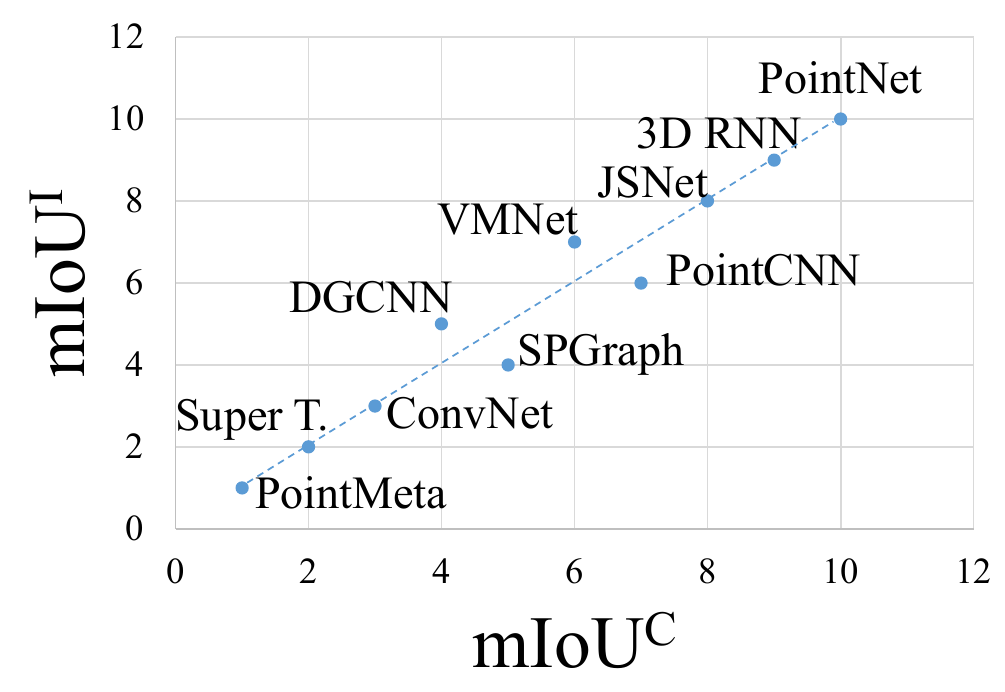}
  \label{fig:subfig:S3DISioui}}
 \end{subfigmatrix}
 \caption{Comparing the rank of $\rm mIoU^{C}$ with $\rm mIoU^{D}$, $\rm mIoU^{P}$ and $\rm mIoU^{I}$ on the S3DIS.}
 \label{fig:rankS3DISiou}
\end{figure}

\begin{figure}[htbp!]
\centering
 \begin{subfigmatrix}{3}                 
  \subfigure[$\rm mIoU^{D}$ and $\rm mIoU^{C}$]{\includegraphics[width=0.31\textwidth]{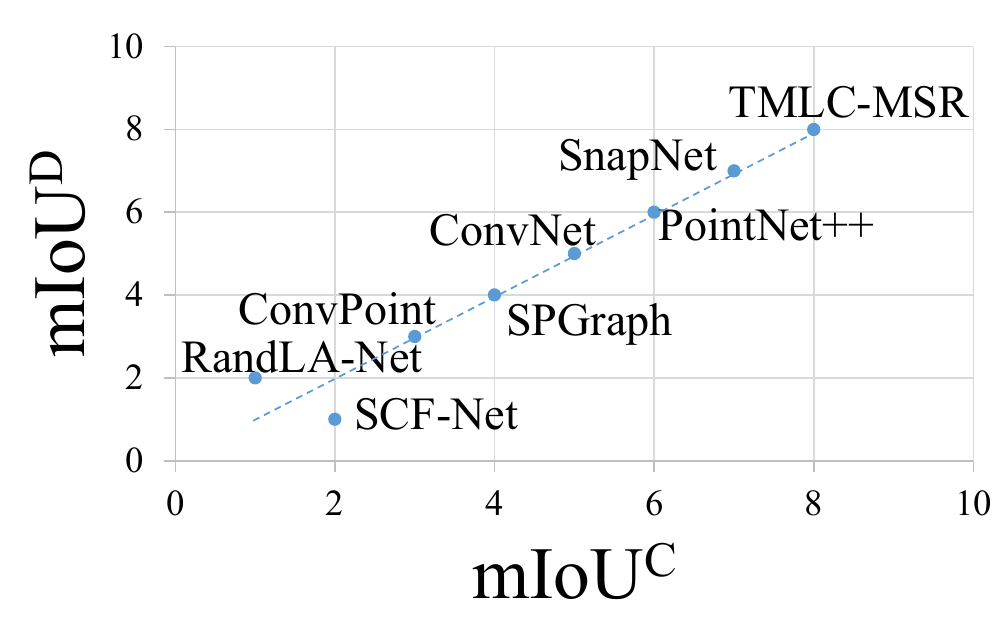}
  \label{fig:subfig:Semantic3Diou}}
  \subfigure[$\rm mIoU^{P}$ and $\rm mIoU^{C}$]{\includegraphics[width=0.31\textwidth]{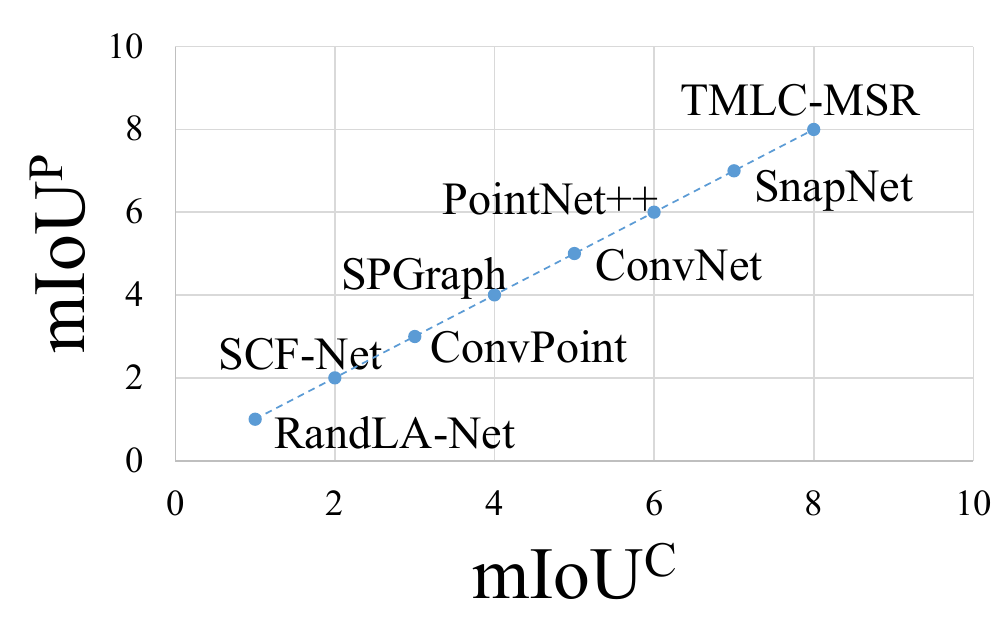}
  \label{fig:subfig:Semantic3Dioup}}
   \subfigure[$\rm mIoU^{I}$ and $\rm mIoU^{C}$]{\includegraphics[width=0.31\textwidth]{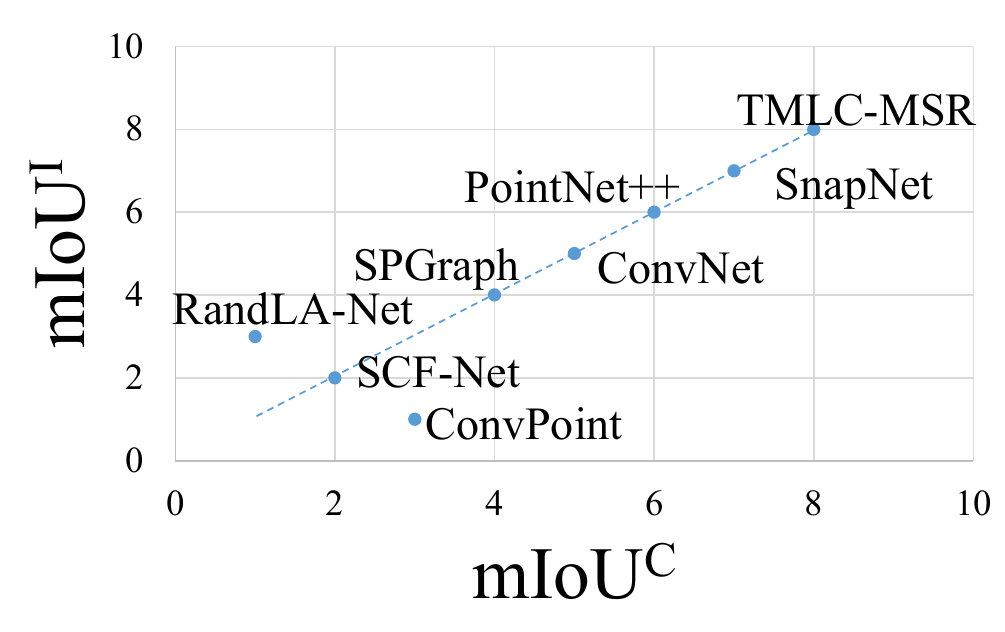}
  \label{fig:subfig:Semantic3Dioui}}
 \end{subfigmatrix}
 \caption{Comparing the rank of $\rm mIoU^{C}$ with $\rm mIoU^{D}$, $\rm mIoU^{P}$ and $\rm mIoU^{I}$ on the Semantic3D.}
 \label{fig:rankSemantic3Diou}
\end{figure}

In addition, this section contrasts the rank of $\rm mIoU^{C}$ with the other three metrics across three datasets, as shown in Figure \ref{fig:rankiou}, \ref{fig:rankS3DISiou} and \ref{fig:rankSemantic3Diou}, respectively.
It is clear that when the performance of multiple methods is close, the experimental results show local ranking discrepancies on several metrics.
Therefore, in order to obtain a more comprehensive evaluation of different segmentation methods, it is required to compare segmentation metrics of varying granularity.

\subsection{Evaluations on mAcc}

This section reports the semantic segmentation results of dataset-level metric $\rm mAcc^{D} $, point cloud-level metrics $\rm mAcc^{P}$ and $\rm mAcc^{C}$, and instance-level metric $\rm mAcc^{I}$ on three datasets, as shown in Table \ref{tab:ScanNetacc}, \ref{tab:S3DISacc} and \ref{tab:Semantic3Dacc}, respectively.
No method achieves the best result across all metrics on the ScanNet and Semantic3D datasets.
On the S3DIS dataset, the PointMeta \cite{lin2023meta} model outperforms other methods on all metrics.
Therefore, it is necessary to simultaneously refer to multiple evaluation metrics of different granularity to carry out comprehensive evaluation.

\begin{table}[htbp!]

	\caption{Comparison of segmentation results for different levels of mAcc metrics on the ScanNet.}
	\centering

	\begin{tabular}{lcccc}
		\hline
		Method & $\rm mAcc^{D} $ (\%)&         $\rm mAcc^{P}(\%) $      &  $\rm mAcc^{C}$ (\%)&  $\rm mAcc^{I}$ (\%)\\ \hline
		PointNet++ \cite{qi2017pointnet++}     &       63.4                             &   70.6 &60.2& 58.7     \\
		PointCNN \cite{li2018pointcnn}   &       71.6                           &   76.1  &69.2&  68.2   \\
		DGCNN \cite{wang2019dynamic}    &          72.3                           &      75.4   &68.8& 66.2\\
		 KPConv \cite{thomas2019kpconv}   &       73.1                             &    78.3 &71.3&  70.2   \\
		SparseConvNet \cite{choy20194d}    &       79.1                              &      85.9  &77.3& 75.8 \\ 
		 VMNet \cite{liu2019densepoint}   &      78.6                                &      85.4  &77.5&  75.9\\ 
   ConvNet+CBL \cite{zhao2021point}   &       81.0                                 &    88.9   &\textbf{79.1}&  \textbf{78.7} \\ 
    PointTransformerV2 \cite{wu2022point}   &       79.7                                &    88.7   & 78.5 & 77.8 \\ 
   OctFormer \cite{wang2023octformer}   &      \textbf{81.3}                                 &    \textbf{90.2}   & 78.9 & 77.7 \\  \hline

	\end{tabular}
	\label{tab:ScanNetacc}
\end{table}

\begin{table}[!h]

	\caption{Comparison of segmentation results for different levels of mAcc metrics on the S3DIS.}
	\centering

	\begin{tabular}{lcccc}
		\hline
		Method & $\rm mAcc^{D} $ (\%)&         $\rm mAcc^{P}(\%) $      &  $\rm mAcc^{C}$ (\%)&  $\rm mAcc^{I}$ (\%)\\ \hline
		PointNet \cite{qi2017pointnet}   &    49.0       &     61.7    &47.6&46.3 \\
		
		PointCNN \cite{li2018pointcnn}   &       63.9                                 &   76.0  &60.8&  59.1   \\
		DGCNN \cite{wang2019dynamic}    &        68.4                                &      82.1 &67.2& 65.8  \\
		SPGraph \cite{2018Large}  &        66.5                                &    81.3  &64.5&  63.1  \\
		3D RNN \cite{ye20183d}   &         65.3                               &      74.2  &60.9& 58.8 \\ 
		 JSNet \cite{zhao2020jsnet}   &        61.4                                &      75.1&59.4& 57.9   \\ 
		 VMNet \cite{liu2019densepoint}    &      63.4                                  &    76.8&61.2&  60.4   \\
		ConvNet+CBL \cite{zhao2021point}   &      75.2                                  &    87.5&73.2&    72.4  \\ 
   PointMeta \cite{lin2023meta}   &           \textbf{86.2}                             &    \textbf{92.9}   &\textbf{83.5}& \textbf{80.2} \\
  Superpoint Transformer \cite{robert2023efficient}   &           85.8                            &    90.6   &81.6&79.1 \\ 
  \hline
		
	\end{tabular}
	\label{tab:S3DISacc}
\end{table}

\begin{table}[!h]

	\caption{Comparison of segmentation results for different levels of mAcc metrics on the Semantic3D.}
	\centering

	\begin{tabular}{lcccc}
		\hline
		Method & $\rm mAcc^{D} $ (\%)&         $\rm mAcc^{P}(\%) $      &  $\rm mAcc^{C}$ (\%)&  $\rm mAcc^{I}$ (\%)\\ \hline
	TMLC-MSR \cite{hackel2016fast}    &     68.2                                &      80.6  &67.8&  66.6\\
		PointNet++ \cite{qi2017pointnet++}  &     71.6                                 &    81.3   &69.8&  68.1 \\
		SnapNet \cite{boulch2018snapnet}   &        75.5                               &      86.2  &72.5&  70.5\\ 
		
		SPGraph \cite{2018Large}    &          81.9                            &    89.5&78.5&77.3\\
		ConvNet+CBL \cite{zhao2021point}    &    82.6                                  &    89.9   &79.6& 77.9 \\ 
    ConvPoint \cite{BOULCH202024}   &       81.5                                 &    90.1   &79.9& 77.7 \\
  RandLA-Net \cite{hu2020randla}   &       82.4                                 &    \textbf{91.6}   &\textbf{81.0}& \textbf{78.6} \\
SCF-Net \cite{fan2021scf}   &      \textbf{82.8}                                 &    91.2   &80.3& 78.0 \\
  \hline
		
	\end{tabular}
	\label{tab:Semantic3Dacc}
\end{table}

\begin{figure}[!h]
\centering
 \begin{subfigmatrix}{3}                 
  \subfigure[$\rm mAcc^{D}$ and $\rm mAcc^{C}$]{\includegraphics[width=0.31\textwidth]{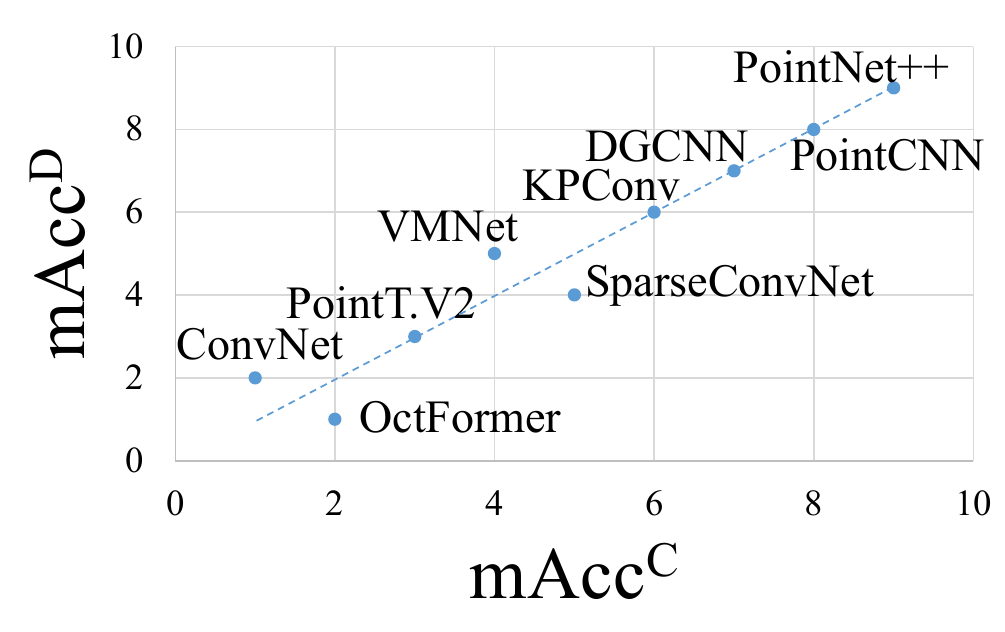}
  \label{fig:subfig:scannetmaccd}}
  \subfigure[$\rm mAcc^{P}$ and $\rm mAcc^{C}$]{\includegraphics[width=0.31\textwidth]{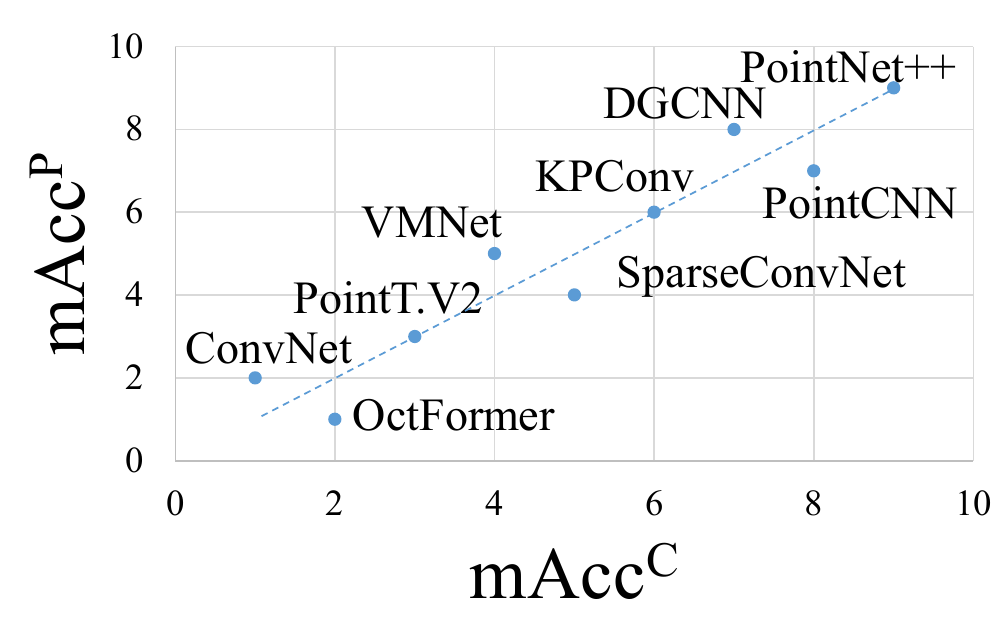}
  \label{fig:subfig:scannetmaccp}}
   \subfigure[$\rm mAcc^{I}$ and $\rm mAcc^{C}$]{\includegraphics[width=0.31\textwidth]{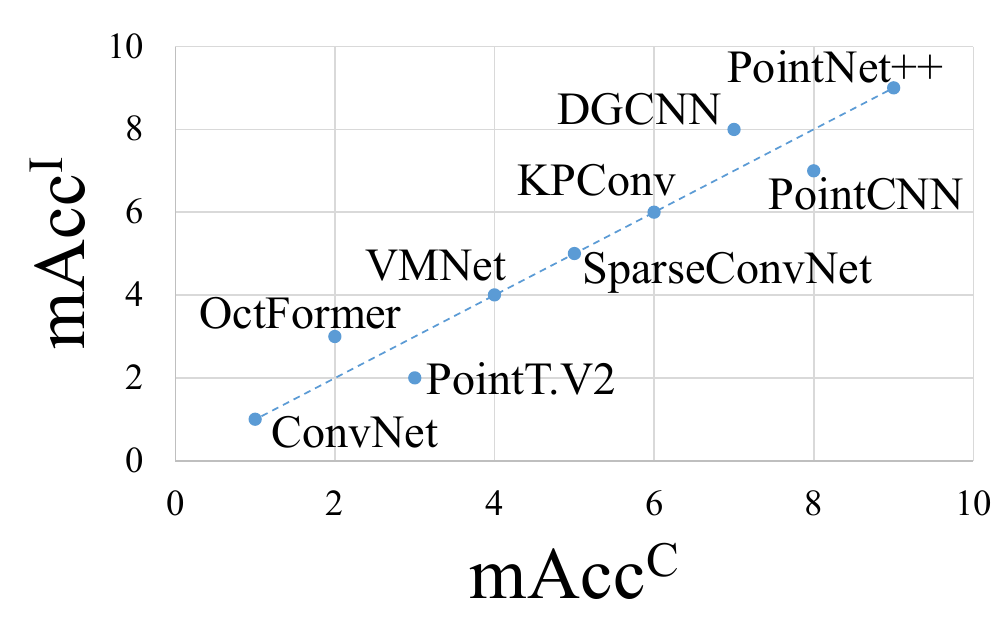}
  \label{fig:subfig:scannetmacci}}
 \end{subfigmatrix}
 \caption{Comparing the rank of $\rm mAcc^{C}$ with $\rm mAcc^{D}$, $\rm mAcc^{P}$ and $\rm mAcc^{I}$ on the ScanNet.}
 \label{fig:rankacc}
\end{figure}

\begin{figure}[htbp!]
\centering
 \begin{subfigmatrix}{3}                 
  \subfigure[$\rm mAcc^{D}$ and $\rm mAcc^{C}$]{\includegraphics[width=0.31\textwidth]{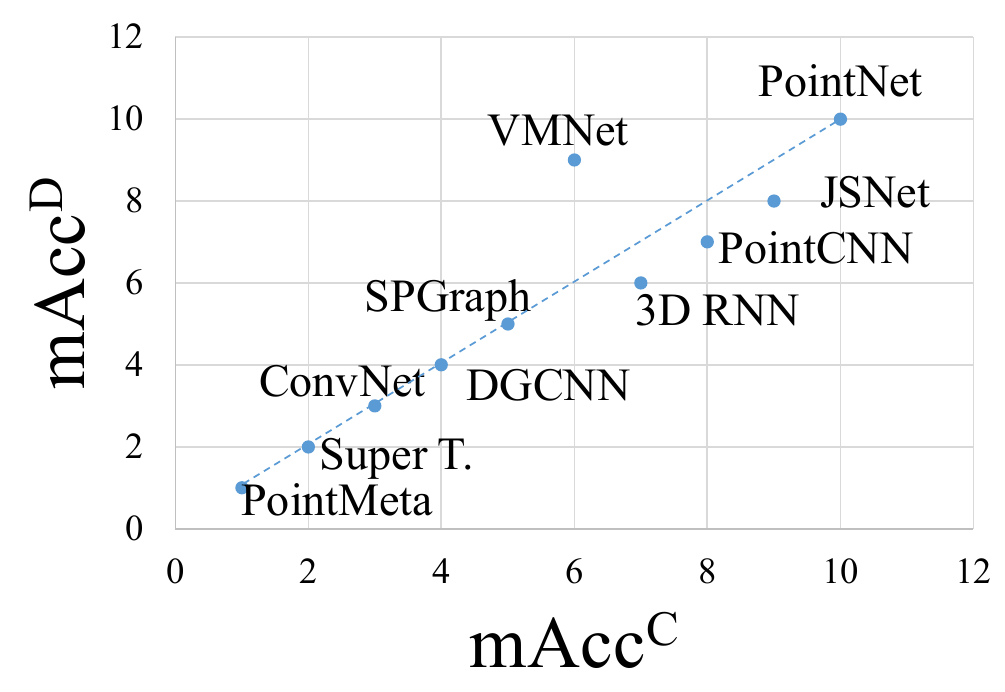}
  \label{fig:subfig:S3DISmaccd}}
  \subfigure[$\rm mAcc^{P}$ and $\rm mAcc^{C}$]{\includegraphics[width=0.31\textwidth]{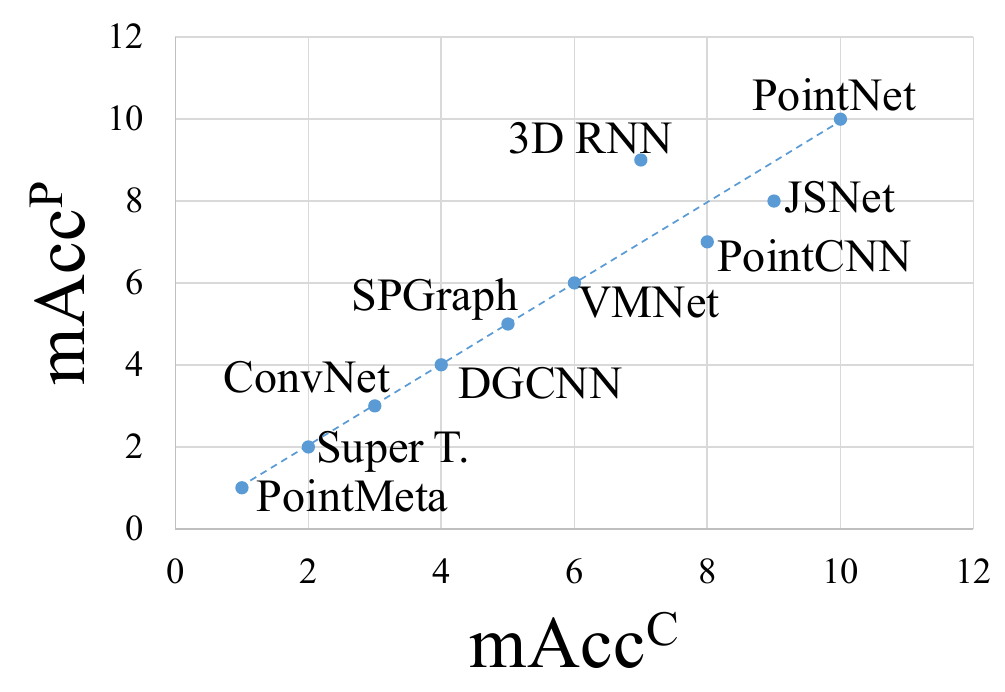}
  \label{fig:subfig:S3DISmaccp}}
   \subfigure[$\rm mAcc^{I}$ and $\rm mAcc^{C}$]{\includegraphics[width=0.31\textwidth]{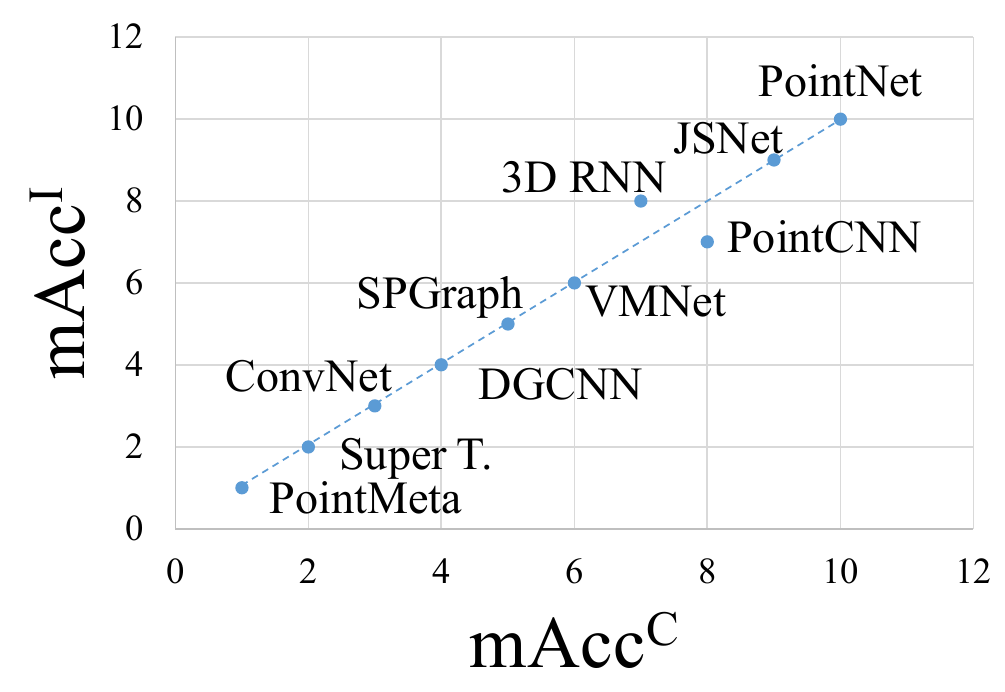}
  \label{fig:subfig:S3DISmacci}}
 \end{subfigmatrix}
 \caption{Comparing the rank of $\rm mAcc^{C}$ with $\rm mAcc^{D}$, $\rm mAcc^{P}$ and $\rm mAcc^{I}$ on the S3DIS.}
 \label{fig:rankS3DISmacc}
\end{figure}

\begin{figure}[htbp!]
\centering
 \begin{subfigmatrix}{3}                 
  \subfigure[ $\rm mAcc^{D}$ and $\rm mAcc^{C}$]{\includegraphics[width=0.31\textwidth]{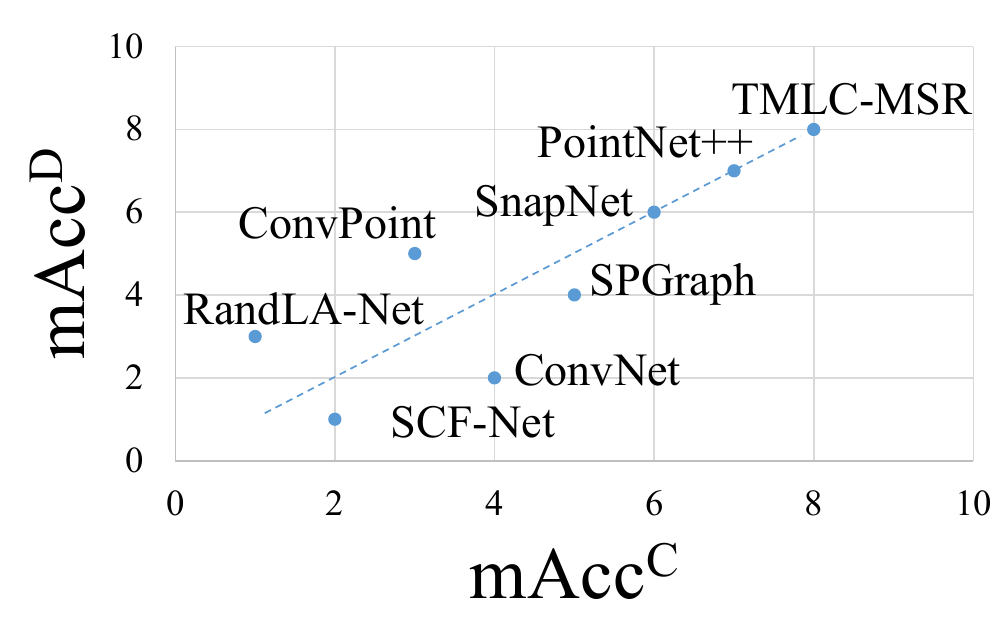}
  \label{fig:subfig:Semantic3Dmaccd}}
  \subfigure[$\rm mAcc^{P}$ and $\rm mAcc^{C}$]{\includegraphics[width=0.31\textwidth]{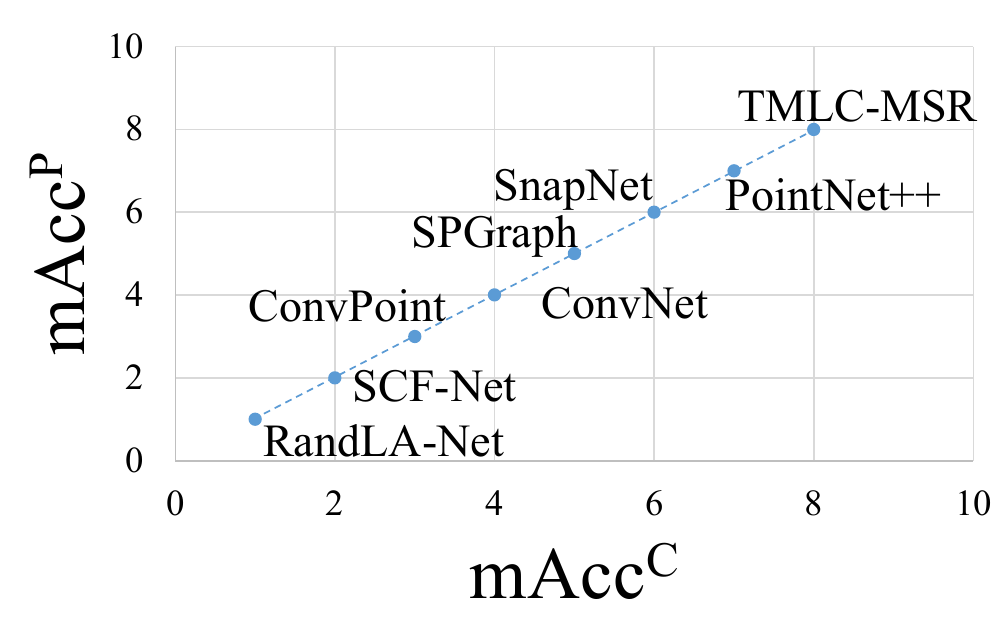}
  \label{fig:subfig:Semantic3Dmaccp}}
   \subfigure[$\rm mAcc^{I}$ and $\rm mAcc^{C}$]{\includegraphics[width=0.31\textwidth]{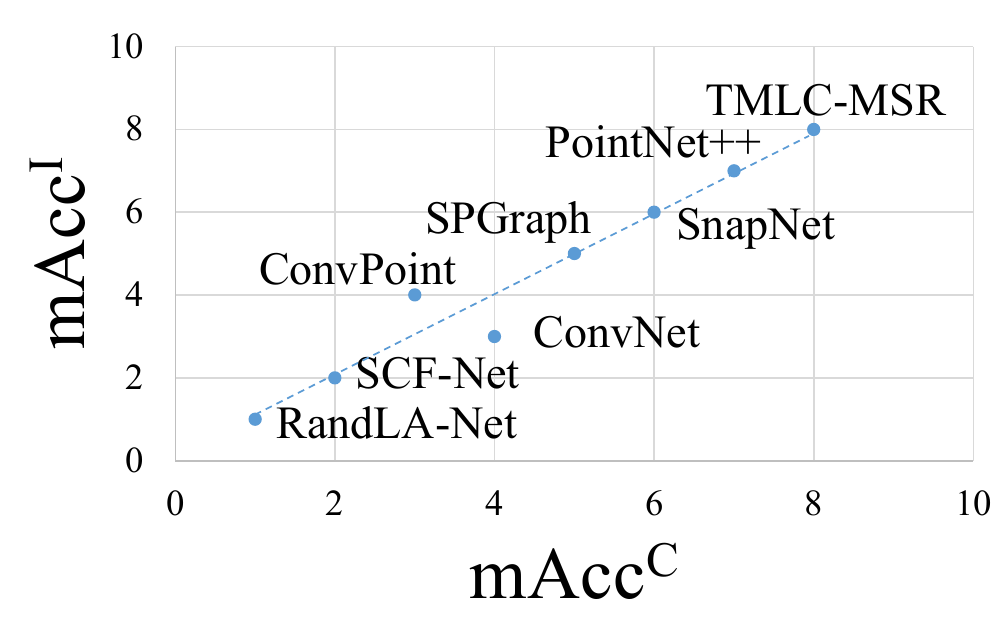}
  \label{fig:subfig:Semantic3Dmacci}}
 \end{subfigmatrix}
 \caption{Comparing the rank of $\rm mAcc^{C}$ with $\rm mAcc^{D}$, $\rm mAcc^{P}$ and $\rm mAcc^{I}$ on the Semantic3D.}
 \label{fig:rankSemantic3Dmacc}
\end{figure}

This section also compares the rank of $\rm mAcc^{C}$ to the other three metrics across three datasets, as illustrated in Figures \ref{fig:rankacc}, \ref{fig:rankS3DISmacc}, and \ref{fig:rankSemantic3Dmacc}, respectively.
Figure \ref{fig:subfig:Semantic3Dmaccp} shows that different methods have consistent rankings between the $\rm mAcc^{P} $ and $\rm mAcc^{C} $ on Semantic3D dataset.
In addition, the segmentation results exhibit local ranking disparities across different metrics when the segmentation performance of different methods is similar.
As a result, the evaluation metrics with different granularity can provide a more comprehensive evaluation.

\section{Conclusion}\label{sec.conclusion}

This paper proposes fine-grained evaluation metrics of semantic segmentation to address the issue that the current evaluation metrics in point cloud semantic segmentation are biased to most categories and large objects.
The suggested fine-grained mIoU and mAcc evaluation metrics at the point cloud and instance levels provide a more thorough evaluation of the segmentation algorithms.
Various semantic segmentation models are trained and assessed using the suggested metrics on three distinct indoor and outdoor semantic segmentation datasets.
Experiments demonstrate that these fine-grained metrics offer greater statistical information and lessen the bias towards large objects.

\section*{Acknowledgment}

The authors sincerely acknowledge the anonymous reviewers for their insights and comments to further improve the quality of the manuscript.
They also would like to thank the participants in the study for their valuable time.

%
%
%
 \bibliographystyle{splncs04}
 \bibliography{samplepaper}
%




\end{document}